\documentclass[runningheads]{llncs}
\usepackage{graphicx}
\usepackage{CJKutf8} 

\begin{document}

\title{XCMRC: Evaluating Cross-lingual Machine Reading Comprehension}
%
%

\author{Pengyuan Liu \inst{1} \and
Yuning Deng \inst{1,2} \and
Chenghao Zhu \inst{1} \and Han Hu\inst{1}}

%
%
\institute{Beijing Language and Culture University \and Tencent Cloud AI
\email{liupengyuan@pku.edu.cn,13552940428@163.com,zhu\_cheng\_hao@163.com,502537185@qq.com}}

\maketitle              

\begin{abstract}
We present XCMRC, the first public cross-lingual language understanding (XLU) benchmark which aims to test machines on their cross-lingual reading comprehension ability. To be specific, XCMRC is a Cross-lingual Cloze-style Machine Reading Comprehension task which requires the reader to fill in a missing word (we additionally provide ten noun candidates) in a sentence written in target language (English / Chinese) by reading a given passage written in source language (Chinese / English). Chinese and English are rich-resource language pairs, in order to study low-resource cross-lingual machine reading comprehension (XMRC), besides defining the common XCMRC task which has no restrictions on use of external language resources, we also define the pseudo low-resource XCMRC task by limiting the language resources to be used. In addition, we provide two baselines for common XCMRC task and two for pseudo XCMRC task respectively. We also provide an upper bound baseline for both tasks. We found that for common XCMRC task, translation-based method and multilingual sentence encoder-based method can obtain reasonable performance but still have much room for improvement. As for pseudo low-resource XCMRC task, due to strict restrictions on the use of language resources, our two approaches are far below the upper bound so there are many challenges ahead.
\end{abstract}

\section{Introduction}

\begin{table*}[ht]
\caption{Samples for XCMRC. For the convenience of presentation, we only present a part of passage from the sample.} 
\footnotesize
\center
\begin{tabular}{|p{2cm}|p{5cm}|p{5cm}|}
\hline
& CPEQ & EPCQ \\
\hline
Passage & \begin{CJK}{UTF8}{gkai}
据周三美国联邦法院听证会的证词显示，在美国国会通过保护职场女性孕期权利的法律三十多年之后，职场女性怀孕依然广泛遭受歧视，需要通过加大宣传和制定更明确的指导方针来和歧视作斗争。

职场女性歧视问题在两周前成为了众人关注的焦点。
\end{CJK}
 & Renowned CCTV anchor Zhao Pu warned of the potential dangers of consuming firm yogurt and jelly in his microblog on Monday .

From the reports , the public widely believed industrial gelatin was being used as an addictive to improve the food 's flavor .
 \\
 \hline
Question & The Pregnancy XXXX Act forbids discrimination by employers based on pregnancy, including hiring, firing, pay, job assignments and promotions.  & \begin{CJK}{UTF8}{gkai}不过这两种XXXX从外观上并无法分辨,给消费者造成困难。\end{CJK}
\\ 
\hline
Candidates & decades, Congress, law, women, workplace, discrimination, publicity, guidelines, testimony,  Wednesday
 &  \begin{CJK}{UTF8}{gkai}食用, 酸奶, 果冻, 京报, 姓名, 明胶, 类食品, 动物, 皮肤, 骨骼\end{CJK}\\
\hline
Answer & discrimination & \begin{CJK}{UTF8}{gkai} 明胶\end{CJK} \\ 
\hline
\end{tabular}
\end{table*} 

A major goal for NLP is to enable machines to understand text to the extent of humans. Several research disciplines are focused on this problem: for example, information extraction, relation extraction, semantic role labeling, and recognizing textual entailment. Yet these techniques are necessarily evaluated individually, rather than by how much they advance us towards the end goal \cite{hermann2015teaching}. In contrast, machine reading comprehension (MRC) is a task where computers are expected to answer question related to a document that they have to comprehend. Such comprehension tests are appealing and challenging because they are objectively gradable and able to measure a range of important abilities, from basic understanding to causal reasoning to inference. Recently, the emergence of a variety of large-scale datasets has fueled up the research phase \cite{D13-1020,hermann2015teaching,C16-1167,W17-2623,kovcisky2018narrativeqa,kwiatkowski2019natural}. Among them, SQuAD \cite{D16-1264} is a typical MRC dataset and has attracted wide attention of academia. 

However, these datasets are all aimed at testing the ability of monolingual understanding and reasoning of machines. This narrows down the application scenarios for MRC systems. In practice, 
existing natural language processing systems used in major international products may need to deal with inputs in many languages. Data annotation requires a lot of efforts so it's unrealistic to annotate all languages a system might encounter during operation. Therefore, cross-lingual language understanding (XLU) has been widely studied. While XLU shows promising results for tasks such as cross-lingual document classification \cite{klementiev2012inducing,L18-1560}, and recently XNLI \cite{D18-1269} is released for cross-lingual natural language inference, there is no any challenging XLU benchmarks for MRC. 

In this work, we introduce a benchmark that we call the Cross-lingual Cloze-style MRC corpus, or XCMRC\footnote{Datasets and codes are
available on https://github.com/NLPBLCU/XCMRC}. It mainly consists of two dual sub-datasets\footnote{We also provide corresponding monolingual MRC sub-dataset. See Section 3.2.}: EPCQ (English Passages, Chinese Questions) and CPEQ (Chinese Passages, English Questions). EPCQ has 57599 samples composed by English passages and Chinese questions while CPEQ has 55990 samples composed by Chinese passages and English questions. Existing XLU benchmarks \cite{klementiev2012inducing,S17-2001,L18-1614} generally consist of train data written in source language and test data written in target language, while EPCQ and CPEQ mix two languages in one data sample as showed in Table 1.

Chinese and English are rich-resource language pairs, so we can define \textbf{the common XCMRC task} which does not have any restrictions on the use of external language resources. For XLU tasks, constructing datasets with low-resource language pairs will be of great significant. But it will take a lot of efforts to build a large-scale one. In order to let our dataset support low-resource language XCMRC research as well, we define \textbf{the pseudo low-resource XCMRC task} which limits model to language resources that most low-resource languages have, such as pre-trained word embeddings. In this way, we can test model aiming at low-resource language XCMRC task on Chinese / English dataset, and that is why we name it as the pseudo low-resource XCMRC task.

We evaluate several approaches on XCMRC. For  pseudo low-resource XCMRC task, we introduce passage independent method which does not use the information of passage, and naive method which employs monolingual MRC model directly. Experimental results show that it is hard to learn enough cross-lingual information by naive method, and it can not reach a good performance depending only on question. For common XCMRC task, translation-based approach which uses a translation system and multilingual-based  approach by fine-tuning multilingual BERT are provided. In addition, we also provide an upper bound baseline \footnote{It is a monolingual MRC model trained on EPEQ and CPCQ, and it could improve along with the performance of monolingual MRC model} for both tasks. We show that though translation-based and multilingual-based approaches can obtain reasonable performance, they still have much room for improvement.

\section{Related Work}  
\subsection{The task of MRC}  

Generally, existing MRC datasets can be categorized into four sub tasks: Extractive MRC\cite{D16-1264,W17-2623,P17-1147,Dunn2017SearchQA,kovcisky2018narrativeqa,P18-2124}, Generative MRC \cite{nguyen2016ms,W18-2605}, Multi-choice MRC \cite{D13-1020,D17-1082} and Cloze-style MRC. 

Cloze-style MRC tasks require the reader to fill in the blank in a sentence. Children’s Book Test (CBT) \cite{Hill2015The} involves predicting a blanked-out word of the 21st sentence while the document is formed by 20 previous consecutive sentences in the book. BT \cite{Hill2015The} is an extension to the named-entity and common-noun part of CBT that increases their size by over 60 times. CNN/Daily Mail \cite{hermann2015teaching} is a dataset constructed from the on-line news articles. This task requires models to identify missing entities from bullet-point summaries of on-line news articles. People Daily  \cite{C16-1167} is the first released Chinese reading comprehension dataset. This dataset is generated automatically by randomly choosing a noun with word frequency greater than two as answer. As we can see, automatically generating large-scale training data for neural network training is essential for reading comprehension.

\begin{table*}[ht]
\caption{Comparison of XCMRC with existing CMRC Datasets} 
\footnotesize
\center
\begin{tabular}{p{2.5cm} p{2.5cm} p{2cm} p{3cm} p{2cm} }
\hline
Dataset & Language & Domain &  Answer type & Provide \\
& & & & candidates\\
\hline
CNN/DailyMail & English & News & Entity & No \\
CBT & English & Children's Book & Noun, named entity, preposition, verb & Yes\\
BT & English & Books & Noun, named entity & Yes\\
People Daily & Chinese & News & Noun & No \\
CFT & Chinese & Children's Fairy Tale & Noun & No \\
\bfseries XCMRC & \bfseries Bilingual & \bfseries News & \bfseries Noun & \bfseries Yes\\
& \bfseries English/Chinese & & & \\
\hline
\end{tabular}
\end{table*} 

\subsection{The task of XLU}  
There have been some efforts on developing cross-lingual language understanding evaluation benchmarks. Klementiev et al.  (2012)  proposed Reuters corpus for cross-lingual document classification \cite{klementiev2012inducing}. Cer et al.  (2017) proposed sentence-level multilingual training and evaluation datasets for semantic textual similarity in four languages \cite{S17-2001}. Agi{\'{c}} and Schluter  (2018) provided a corpus consisting of human translations for 1332 pairs of the SNLI data into Arabic, French, Russian, and Spanish \cite{L18-1614}. Conneau et al. (2018) proposed cross-lingual natural language inference corpus benchmark (XNLI) which consists of 7500 human-annotated development and test examples in NLI three-way classification format in 15 languages \cite{D18-1269}. Cross-lingual question answering (XQA) has been widely studied \cite{Bouma2008Question,Mitamura2010Overview,Soboroff2016The,Ture2016Learning,pouranbenveyseh:2016:TextGraphs-10}. Joty et al. (2017) presented a cross-lingual setting for community question answering \cite{Joty2017Cross}.

\section{The XCMRC Task}

\subsection{Task Definition}

The XCMRC sample can be formulated as a quadruple: $\langle D, Q, A, C  \rangle $, where D is the document or passage, Q is the query, A is the answer to the query and $C$ denotes the candidates. The question $Q$, the answer $A$ and the candidates $C$ are written in target language while the document $D$ is written in source language. 
The XCMRC task requires a model read the document written in source language and then answer the question written in target language. Specifically, the model is required to choose a word from the candidates $C$ and then fill in the blank of the question $Q$ after reading the document $D$. 

\subsection{The XCMRC Corpus}
As mentioned above, the XCMRC corpus mainly consists of two sub-datasets: EPCQ and CPEQ. In order to set up a reasonable upper bound for our task, we additionally construct two corresponding monolingual MRC datasets: EPEQ (English  Passages, English Questions) and CPCQ (Chinese  Passages, Chinese Questions). In this section, we will describe the construction process in detail.

\subsubsection{The Bilingual Corpus}
 We have collected a raw bilingual parallel corpus from a high-quality English language learning website (The Economist channel) \footnote {http://www.kekenet.com/Article/media/economist/}. The corpus consists of 25467 bilingual articles. These articles cover a wide range of topics, from financial to education to sports. Each bilingual article is composed by a set of Chinese paragraphs $paraC=\{{pc}_{1}, {pc}_{2}, \cdots ,{pc}_{m}\}$ and a set of responding English paragraphs $paraE=\{{pe}_{1},{pe}_{2}, \cdots ,{pe}_{m}\}$. $pc$ denotes the paragraph written in Chinese and $pe$ denotes the paragraph written in English. Paragraphs are strictly aligned.

We do part-of-speech tagging for the Chinese paragraphs using Jieba\footnote{https://pypi.org/project/jieba/} and the English paragraphs using NLTK\footnote{http://www.nltk.org/}. 


\subsubsection{Automatic Generation of EPCQ and CPEQ Datasets}
The detailed generating procedures are as follows.
\begin{itemize}
\item We count the frequency for all the nouns appearing in the Chinese passage $paraC$ and thus form a noun set ${C}^{'}$. We choose nouns\footnote{Nouns with frequency between 3 and 10. We count and get the frequency distribution of nouns in Chinese passage and the moderate interval (3-10) was selected.} from ${C}^{'}$ to form the answer candidate set ${A}^{'}$. 
\item Randomly choose an answer word $A$ from the answer candidate set ${A}^{'}$. When chosen, the answer word $A$ will be deleted from set ${A}^{'}$. Find all paragraphs from $paraC$ which contain the answer word ${A}$ and thus form the question candidate set ${Q}^{'}$. Then randomly choose a paragraph from ${Q}^{'}$ with sequence length greater than 10 to generate the question. The question is formed by replacing the answer word $A$ with a placeholder ``XXXX". If the answer word $A$ appears many times in this chosen paragraph, only the first position answer word appearing in the paragraph will be replaced. The corresponding English paragraph ${{pe}_{j}}$ is removed from the ${paraE}$ and the remaining paragraphs of ${paraE}$ $\{{pe}_{1},\cdots,{pe}_{j-1},{pe}_{j+1}, \cdots ,{pe}_{m}\}$ will be used to form ${D}$.
\item We randomly choose nine nouns from the noun set ${C}^{'}$. The nine incorrect answer words and the answer word $A$ together form the candidate set $C$. Thus there are ten nouns in the final candidate set ${C}$.
\item The tuple $\langle D, Q, A, C\rangle $ forms a sample.
\end{itemize}

The above version is referred as EPCQ, and CPEQ is generated in the similar way through interchanging English / Chinese.
\begin{table}[ht]
\caption{Data comparison with existing CMRC datasets. Statistics are taken from \cite{hermann2015teaching,C16-1167}} 
\small
\center
\begin{tabular}{c c c c}
\hline
Dataset & Train set & Test set & Dev set\\
\hline
CNN & 380,298 & 3,924 & 3,198\\
DailyMail & 879,450 & 64,835 & 53,182\\
CBT Common Nouns & 120,769 & 2000 & 2500 \\
CBT Named Entities & 108,719 & 2000 & 2500 \\
BT & 14, 140, 825 & 10,000 & 10,000 \\
People Daily & 870,710 & 3000 & 3000 \\
Children's Fairy Tale & 0 & 3599 & 0 \\
\bfseries EPCQ / EPEQ & \bfseries 54,599 & \bfseries 1500 & \bfseries 1500 \\
\bfseries CPEQ / CPCQ & \bfseries 52,990 & \bfseries 1500 & \bfseries 1500 \\
\hline
\end{tabular}
\end{table}

\begin{table*}[ht]
\caption{Statistics for the XCMRC corpus} 
\small
\center
\begin{tabular}{p{4cm} p{1cm} p{1cm} p{1cm} p{1cm} p{1cm} p{1cm}}
\hline
&& EPCQ & & & CPEQ & \\
& Train & Dev & Test & Train & Dev & Test \\
\hline
Avg \# document length & 544 & 530 & 385 & 536 & 510 & 382 \\
Avg \# question length & 53 & 55 & 42 & 55 & 56 & 44\\
Max \# document length & 8786 & 3584 & 2683 & 8381 & 3366 & 2733 \\
Max \# question length & 463 & 225 & 172 & 468 & 323 & 195 \\
\hline
\end{tabular}
\end{table*} 
\subsubsection{Corresponding Monolingual MRC Sub-datasets: EPEQ and CPCQ}
 These two sub-datasets are constructed in the same way as EPCQ and CPEQ except that the document ${D}$ is formed directly using the paragraphs written in the same language as the question ${Q}$. That is, if we choose a paragraph from $paraE$ as the question, then the remaining paragraphs of $paraE$ will be used as the document. This means, EPEQ is an English cloze-style dataset similar to CBT \cite{Hill2015The}, and CPCQ is a Chinese cloze-style dataset similar to People Daily \cite{C16-1167}.

\subsubsection{The Resulting Dataset}\footnote{Because EPEQ and CPCQ are the corresponding monolingual MRC datasets, the statistics is basically same to EPCQ and CPEQ. Later, we will mainly describe EPCQ and CPEQ in detail.}
Finally, we have generated 57599 samples for EPCQ / EPEQ and 55990 samples for CPEQ / CPCQ . Samples for our dataset are shown in Table 1. Comparison between XCMRC and existing cloze-style datasets are shown in Tabel 2 and Table 3. Specific statistical information for XCMRC is listed in Table 4.

\section{Approaches for XCMRC}

\subsection{Translation-Based Approaches}
For common XCMRC task, the most straightforward techniques rely on translation by which turns XCMRC task into monolingual MRC task. There are two common ways to use a translation system: TRANSLATE QUESTION, where the question and ten candidates of a sample are translated into source language; TRANSLATE PASSAGE, where the passage of a sample is translated into target language. Both approaches are limited by the quality of the translation system, especially the former. Because it needs to translate ten context-less candidates correctly, which are all single words. Using translation system to translate single word is very difficult, because of the polysemy phenomenon in human language whereas focusing on the word translation is not the original intention of XCMRC. Thus we only use the TRANSLATE PASSAGE way as the baseline of translation-based approaches in this paper.

There are a lot of models for monolingual MRC and we choose BiDAF \cite{DBLP:journals/corr/SeoKFH16} which is a popular and high-performance one as our prototype model. The original BiDAF chooses answer word from document and we slightly change it to force the model to choose answer from both the document and the candidates. 

We introduce \textbf{BiDAF\_Cloze}. We compute a score for each word in the context as the probability indicates whether it is the right answer. An extra answer mask is added to force model to choose answer from the candidates. This model has changed the modeling layer and output layer of BiDAF as follows:   

\begin{equation}
output = \mathop{\mathbf{softmax}}_{a \in C \cap D }\ ({{W}_{A}G}),\ {W}_{A} \in R^{1 \times 8d}
\end{equation}

Here $C$ denotes the candidates, $D$ denotes the document, $G$ denotes the output of Attention Flow Layer of BiDAF. $d$ is the dimension of word embedding.

\subsection{Naive Approaches}
It is a natural and worth-trying idea to use ready-made monolingual MRC methods on XCMRC directly. We call it naive approaches and take it as a baseline for low-resource XCMRC task. For better comparison with translation-based models, we still choose BiDAF as a prototype model here. For XCMRC, document and question are written in different languages so that the model cannot be designed to choose answer from the document. BiDAF \cite{DBLP:journals/corr/SeoKFH16} is also designed to extract answer from document, so we need to revise the answer layer of it to adapt to our task. 

We introduce \textbf{BiDAF\_Candidates}. This model has changed the modeling layer and output layer of BiDAF as follows:

\begin{equation}
k  = \mathbf{softmax}(W_{g}G), G_{1} = \sum_{j=1}^m{k_{j}}{{G_{:j}}}
\end{equation}

\begin{equation}
\tilde{A} = HighwayNetwork(A)
\end{equation}

\begin{equation}
output = \mathbf{softmax}(G_{1}W_{B}\tilde{A})
\end{equation}

\begin{equation}
G_{1} \in R^{8d}, A \in R^{10 \times d}, W_B \in R^{8d \times d}
\end{equation}

Here ${A}$ denotes the word embedding matrix  for candidates $C$.
\subsection{Passage Independent Approaches}
Suppose you can not understand Chinese passage, could you choose the right answer only use the information from the English question itself?
Although sometimes the information within the question is not adequate, the above methods can work under certain circumstances. For example, humans can easily choose the answer (``\emph{discrimination}") given the question (``\emph{The Pregnancy XXXX Act forbids discrimination by employers based on pregnancy, including hiring, firing, pay, job assignments and promotions}") with ten candidates (``\emph{decades, Congress, law, women, workplace, discrimination, publicity, guidelines, testimony,  Wednesday}") without reading the passage. We introduce \textbf{PI\_Candidates} (Passage Independent), to study to what extent a model can solve XCMRC task only use question information. We generate a passage-independent representation $Q^{indep}$ for question and then use it to interact with the ten candidates.
\begin{equation}
l  = \mathbf{softmax}(W_{q}U),\ \ \ W_{q} \in R^{1 \times 2d}
\end{equation}
\begin{equation}
Q^{indep} = \sum_{j=1}^n {l_{j}}{{U_{:j}}},\ \ \ Q^{indep} \in R^{2d}
\end{equation}
\begin{equation}
output = \mathbf{softmax}(Q^{indep}W_{C}\tilde{A}),W_C \in R^{2d \times d}
\end{equation}    

Here $U$ denotes the output of Contextual Embedding Layer for question $Q$ of BiDAF. 
\subsection{Multilingual Sentence Encoder-Based Approach (MSE-based)}

Instead of translating the document into target language, we can use a multilingual sentence encoder to represent it and then narrow down the language barrier. This type of method works for common XCMRC of which multilingual sentence encoder is easily obtained because there are plenty of parallel corpus.

There has been some efforts on developing multilingual sentence embeddings.
Zhou (2016) learned bilingual document representations by minimizing the Euclidean distance between document representations and their translations \cite{Zhou2016CrossLingualSC}.
Conneau (2017) and Espa (2017) jointly trained a sequence to sequence MT system on multiple languages to learn a shared multilingual sentence embedding space \cite{Conneau2017Supervised} \cite{Espa2017An}. Our method leverages the latest breakthrough in NLP: BERT \cite{Devlin2018BERT} as the multilingual sentence encoder. BERT has been proved as an effective sentence encoder in many NLP tasks and gained a lot of attention. 

We introduce \textbf{BERT\_Candidates}, which is a combination of a multilingual version of BERT and BiDAF\_Candidates. The multilingual version of BERT model provided by Google\footnote{https://github.com/google-research/bert} uses character-based tokenization for Chinese. Since the passages in XCMRC corpus are very long, if we tokenize the Chinese passage into lists of characters, the vector representation for passage will take up a lot of GPU RAM. Intuitively, using pre-trained word embeddings for Chinese words will be more effective because the answer word is an single word. So we only train BERT\_Candidates on EPCQ and use BERT to get the contextual representation for English passage. As for Chinese words, we use pre-trained word embeddings to represent them. The other components of  BERT\_Candidates are the same with BiDAF\_Candidates.
\begin{equation}
{H} = BERT(P_{BERT})
\end{equation}

Here $P_{BERT}$ indicates the word token ids created from the vocabulary of pre-trained BERT model. ${H}$ works the same as the output of Contextual Embedding Layer of BiDAF for document.

\section{Experiments and Discussion}

\subsection{Experimental setup}
For the translation-based approach, we use Baidu Translation API \footnote{http://api.fanyi.baidu.com/api/trans/product/index/} to translate the document of the dev set.

We count the frequency of the whole XCMRC corpus(including train set, dev set and test set) and keep the top 95\% words as our vocabulary. We use 300D pre-trained word embeddings trained by glove\footnote{ http://nlp.stanford.edu/projects/glove/, for English. https://github.com/embedding/chinese-word-vectors/, for Chinese.} for initialization. As for BERT\_Candidates model,  we use the vocabulary table provided by the multilingual version of BERT model for English and our own vocabulary for Chinese words. We use Tensorflow to complete our models. As for BERT\_Candidates model, we use the Adam optimizer with learning rate 0.0001. For other models, we use the Adam optimizer with learning rate 0.001. We sort all the examples by the length of its document, and randomly sample
a mini-batch of size 25 for each update\footnote{Note that for BERT\_Candidates model, we set the batch-size to 6.}. We trained model for 10 epochs and choose the best model according to the performance of dev set. We run our models 5 times independently with the same
random seed 1234 and report average performance
across the runs.

\subsection{Results and Analysis}

\begin{table*}[ht]
\caption{Baselines on XCMRC. }
\footnotesize
\center
\begin{tabular}{ p{3.5cm}  p{1.2cm} p{1.2cm} p{1.2cm} p{3cm} p{2.5cm}}
\hline
Baselines & Passage & Question & & Task & Model \\

\indent & \indent & English & Chinese & \indent\\
\hline
Naive & English & 58.35\% & 59.43\%  & pseudo low-resource & BiDAF\_candidate\\
\indent & Chinese &  61.64\% & 59.83\%  & XCMRC \\
\hline 
Passage Independent & English & 59.83\% & 59.83\%  & pseudo low-resource  & PI\_candidate\\
\indent & Chinese & 58.20\% & 58.20\%  &XCMRC\\
\hline
Translation-based & English & N/A & 65.99\% & common XCMRC  & BiDAF\_Cloze\\
\indent & Chinese & 67.28\% & N/A & \indent\\
\hline 
MSE-based & English & N/A & 63.28\%  & common XCMRC  & BERT\_Candidates\\
\hline
Upper Bound & English & 72.97\% & N/A  & both of XCMRC task  & BiDAF\_Cloze\\
\indent  & Chinese & N/A & 68.81\% & & \indent\\
\hline 

\end{tabular}
\end{table*}

We evaluate the models in terms of accuracy. For the convenience of presentation, we only present the performance on dev set. The overall experimental results are represented in Table 5. It shows the upper bound of XCMRC has reached 72.97\% and 68.81\% for CPEQ and EPCQ respectively.The performance on EPCQ is a bit lower than CPEQ. Note that this upper bound would keep increasing along with the promotion of the performance on monolingual MRC model. For example, after BiDAF reached 77.3\% F1 score on SQuAD v1.1, the best F1 scores evaluated on the test set of SQuAD v1.1 is 93.16\footnote{https://rajpurkar.github.io/SQuAD-explorer/} now. We expect the newest model on our task would improve the upper bound significantly. Of course, the performances of all the models exceed the random choice (10\%) by a big margin. It means that all models can learn information that is helpful for XCMRC task to a certain extent. It's not surprising that the performance of each model on low-resource XCMRC task is much lower than that of common XCMRC task. The average performance margin between low-resource XCMRC task and common XCMRC task is within 7\%.

For low-resource XCMRC task, the average performance between the naive approaches and the passage independent approaches is relatively close. This may be due to the fact that naive approach, which learns cross-lingual information directly, has learned very limited information about the document. So its performance is comparable to passage independent approaches which only utilize question information. We have noticed that for naive approach, the performance of CPEQ (61.64\%) was about 3\% higher than that of EPEQ (58.35\%). We can not explain it from our experiences. We guess that it's because BiDAF\_Candidates cannot utilize the contextual information of the document effectively as BiDAF\_Cloze does and thus lead to accidental performance. So there are many challenges ahead for the pseudo low-resource XCMRC task.



For common XCMRC task, translation-based approach obtain the best performance (67.28\%, 65.99\%) and still have room for improvement. The results are much like that of XNLI \cite{D18-1269} in which translation-based methods are the best too.

\section{Conclusion}

There has been a growing interest in cross-lingual understanding, since the lack of supervised data for languages in industrial application and annotating data in every language is not really realistic. In this work, we introduce a public XLU benchmark which aims to test machines on their XMRC ability. The dataset, dubbed XCMRC, is the first cross-lingual cloze-style machine reading comprehension dataset. Meanwhile, besides common XCMRC task, we also define the pseudo low-resource XCMRC task in order to support XLU research of low-resource languages. We present several approaches as the baselines of XCMRC. We found that there are many challenges ahead for pseudo low-resource XCMRC task, both passage independent approach and naive approach can not learn enough cross-lingual information. And it is indeed too difficult to learning cross-lingual information under the strict restrictions of low-resource XCMRC. If we loosen up restrictions a little, for example, allowing use a small parallel dictionary or a small-scale parallel corpus, multilingual word embeddings could be a worth trying way. While for common XCMRC task, translation-based method obtains the best performance but it relies on translation system excessively. The multilingual sentence representation model provides reasonable performance, and we think it is a promising research approach in future work. XCMRC opens up several interesting research avenues to explore novel neural approaches for studying XLU ability.

\section{Acknowledgements}
This work is supported by Beijing Natural Science Foundation(4192057).

\bibliographystyle{splncs04}
\bibliography{papers.bib}
%





\end{document}